\documentclass[twoside]{article}

\usepackage{aistats2020}
\usepackage{graphicx}
\usepackage{bm}
\usepackage[ruled,vlined]{algorithm2e}
\usepackage{booktabs} 
\usepackage{amsmath}
\usepackage{comment}
\usepackage[font=small,skip=0pt]{caption}
\usepackage[utf8]{inputenc}
\usepackage{csquotes}
\usepackage{dirtytalk}
\usepackage{epigraph}
\usepackage{color}
\usepackage{amssymb}
\usepackage{breakcites}
\usepackage{multirow}
\usepackage{tabularx}

\begin{document}

%

%

\twocolumn[

\aistatstitle{Supervised Feature Subset Selection and Feature Ranking for Multivariate Time Series without Feature Extraction}

\aistatsauthor{ Shuchu Han \And Alexandru Niculescu-Mizil }
\aistatsaddress{ NEC Laboratories America } ]

\begin{abstract}
We introduce supervised feature ranking and feature subset selection algorithms for multivariate time series (MTS) classification. Unlike most existing supervised/unsupervised feature selection algorithms for MTS our techniques do not require a feature extraction step to generate a one-dimensional feature vector from the time series. Instead it is based on directly computing similarity between individual time series and assessing how well the resulting cluster structure matches the labels. The techniques are amenable to heterogeneous MTS data, where the time series measurements may have different sampling resolutions, and to multi-modal data.

\end{abstract}

\section{Introduction}
From cyber-physical systems to IoT to healthcare, multi-dimensional time series data is ubiquitous in many applications and it is collected at an increasing rate and by an increasing number of sensors.  It is not uncommon for larger systems to be instrumented with thousands of sensors making multiple measurements a second. The transmission, storage, processing and analysis of such a large  amount of data is often impractical, especially if the analysis must be real-time, or must be performed on low powered edge computing devices. To cope with this problem practical systems often use only data from a small subset of informative sensors, while irrelevant or redundant sensors are ignored.

\begin{figure*}[ht!]
\centering
\begin{minipage}[l]{0.9\linewidth}%
\centering
\includegraphics[width=0.55\linewidth]{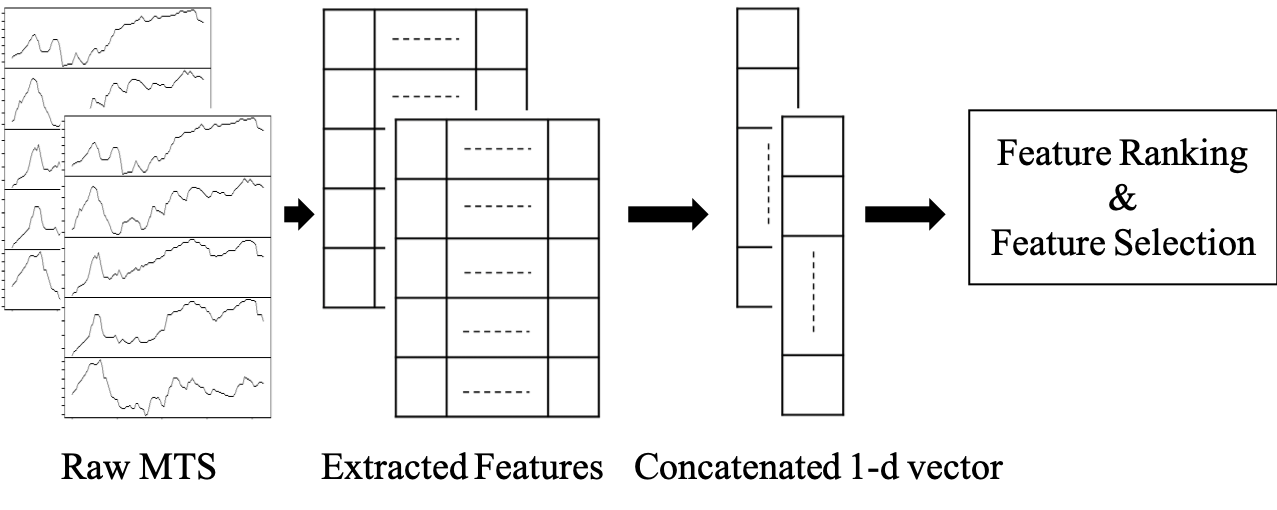}
\end{minipage}\\%
\begin{minipage}[l]{0.9\linewidth}%
\centering
\includegraphics[width=0.8\linewidth]{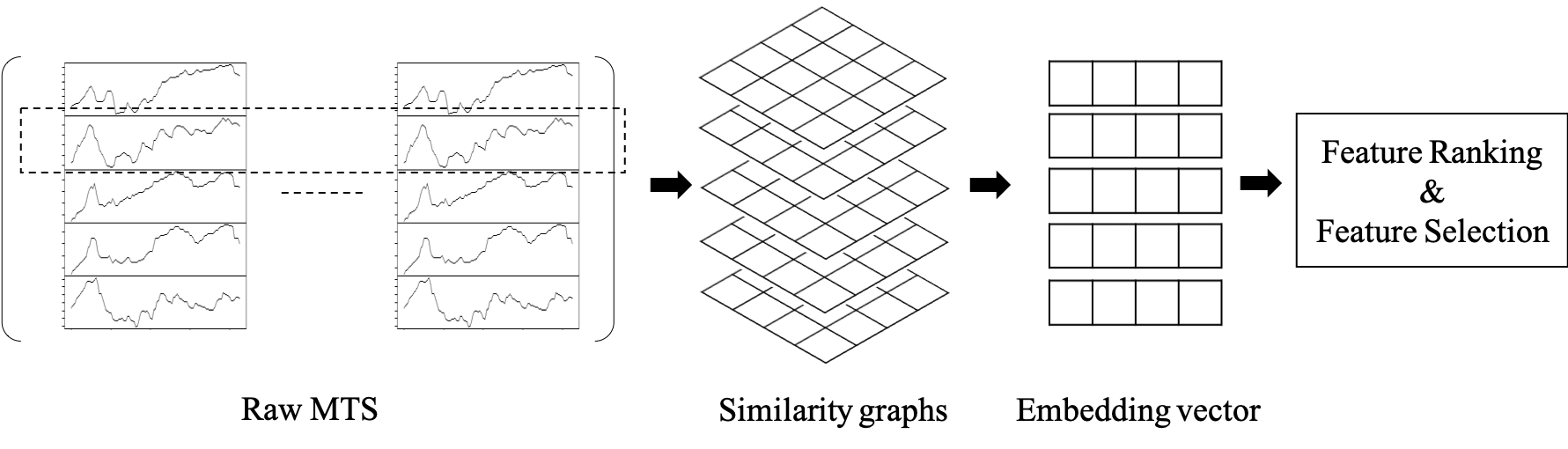}
\end{minipage}%
\caption{Top: common approach for feature selection based on feature engineering. Bottom: proposed solution in this work.}%
\label{fig:teaser}%
\end{figure*}

In this paper we tackle the sensor selection problem in the context of multivariate time-series classification where the task is to assign one label to an entire MTS segment. The traditional approach to this problem is vectorize the data by  extracting several features from each time-series in an MTS segment (e.g. mean value, standard deviation, spectrum, etc. ) and  concatenate the features from all time-series into an 1-D feature vector of fixed dimension. Once this vectorized representation of an MTS segment is obtained, classical feature selection can be used to obtain a small subset of informative features (see Figure~\ref{fig:teaser}). One downside of this approach is that it is critically reliant on feature engineering and on the user's domain knowledge to decide what kinds of features to extract. For example, one we may say that the spectrum of sound at different frequency is a deciding factor for classifying different pronunciations; or the mean value of stocks is one of the crucial factors for the value of S\&P 500 index. This, however, may be difficult when the user has little knowledge or intuition on the nature of the connection between the different time-series and the label. In this case one would have to resort to simply extracting well-known time-series features using some tool such as TSFRESH~\cite{li2016feature} and hope for the best. 

To circumvent the feature engineering problem we propose two supervised sensor selection algorithms for multivariate time-series classification that do not require the vectorization of the MTS segments. Instead, we only require a distance measure between time-series which can be calculated directly using, for example, Dynamic Time Warping (DTW) \cite{berndt1994using}.  The key intuition behind our algorithms is that, if a sensors produces similar time-series in segments with the same label, and dissimilar time-series in segments with different labels, then that sensor is likely an informative one. 

Based on this idea, we propose both a sensor ranking algorithm (akin to filter based methods in classical feature selection) and a sensor subset selection algorithm. For sensor ranking, we calculate a relevance score for each sensor by first constructing a similarity graph among the time-series produced by the sensor across all MTS segments. We then find the largest eigenvector of the normalized adjacency matrix of this graph, which reflects its  cluster structure \cite{shi2000normalized}. Finally, the relevance score of a sensor is calculated as the normalized mutual information between this eigenvector and the ground truth labels (Figure~\ref{fig:teaser}).  

While the sensor ranking technique is simple and efficient, it will assign a similar relevance score to highly correlated sensors thus potentially leading to the selection of redundant sensors. To address this problem we also propose a sensor subset selection algorithm. We again start by calculating an adjacency matrix for each sensor, then find a linear combination these matrices that (1)approximates similarity matrix of the labels and (2)uses a small number of sensors and (3)uses minimally redundant sensors. 
 
Since the proposed techniques are only based on distance measures between time-series produced by the same sensor do not require that all sensors sample data at the same rate. This makes them readily applicable to heterogeneous MTS data where different sensors have different sampling rates, without needing to sub-sample high frequency sensors or interpolate low frequency sensors in order to convert the MTS data to a matrix format. This is a major advantage over MTS sensor selection techniques based on inter-sensor correlation such as CLeVer \cite{yoon2005feature} or Corona \cite{yang2005supervised} which require all sensors to have the same sampling rates.

The rest of paper is organized as follows. In Section \eqref{sec:math_notation}, we provide some background and the math notations that are used in our equations. In Section~\eqref{sec:algorithm}, we present our algorithms for sensor ranking and sensor subset selection. In section~\eqref{sec:discussion}, we discuss our connection to two-stage kernel learning algorithms. The experiments and discussion are introduced in Section~\eqref{sec:experiments}. In Section~\eqref{sec:heterogeneous}, we extend our work to the application on heterogeneous data. And we summarize our work in Section~\eqref{sec:conclusion}.

\begin{figure*}[h!]
\centering
\begin{minipage}[l]{\linewidth}%
\centering
\includegraphics[width=0.9\linewidth]{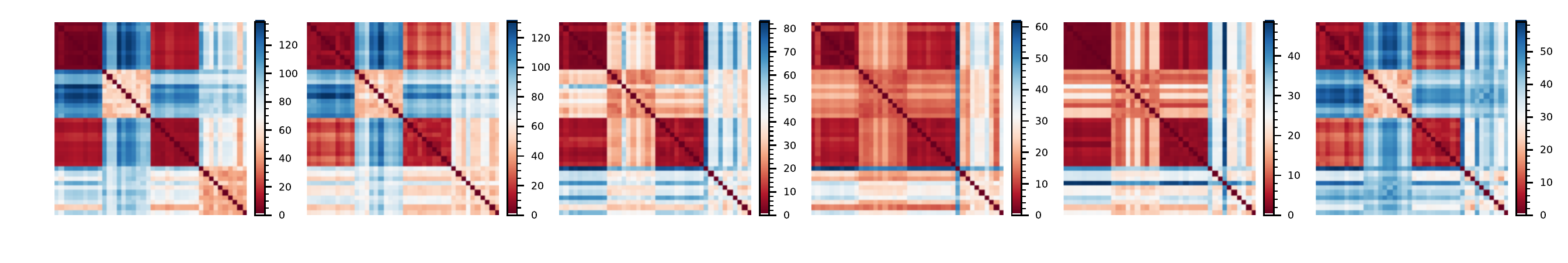}
\end{minipage}\\%
\begin{minipage}[l]{\linewidth}%
\centering
\includegraphics[width=0.9\linewidth]{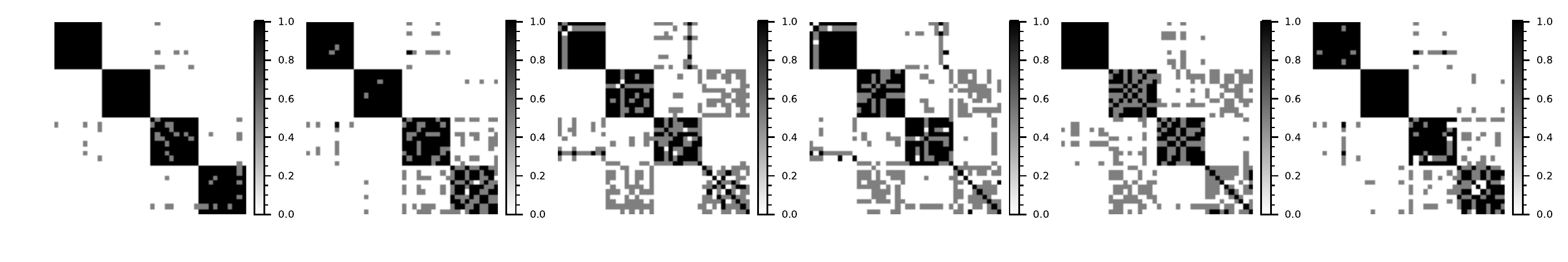}
\end{minipage}\\%
\begin{minipage}[l]{\linewidth}%
\centering
\includegraphics[width=0.9\linewidth]{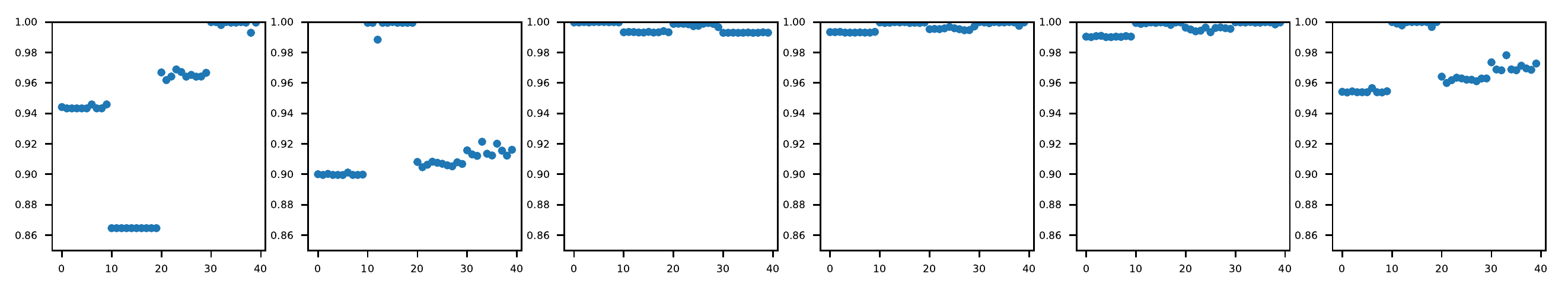}
\end{minipage}%
\caption{Illustration of our idea by using the ``BasicMotions" dataset. The MTS samples are sorted according to their labels (``Standing", ``running", ``walking", ``badminton") from left to right as in each subplots on purpose. The goal is to observe the cluster structure revealed by the DTW distance. Top: visualization of distance matrix where the distances is calculated by DTW. Center: Affinity matrices of constructed $k$-nearest neighbor graph. Bottom: the corresponding spectral embedding vector by PIE.}%
\label{fig:MTS_PIE}%
\end{figure*}

\section{Background and Notations}
\label{sec:math_notation}

Assume we are given a labeled data set $\{(\bm{X}_i, y_i)\}_{i=1..n}$, where $\bm{X}_i$ is a MTS segment, and $y_i \in \mathcal{R}$ is the corresponding label. The each MTS segment $\bm{X}_i$ consists of $m$ time-series or sensors $\{\bm{x_{i,j}}\}_{j=1..m}$ with $\bm{x_{i,j}} \in \mathcal{R}^{l_{i,j}}$. The length $l_{i,j}$ of each time-series $\bm{x_{i,j}}$ may vary between segments due to different segment duration (i.e. $l_{i_1,j} \neq l_{i_2,j}$) or between sensors due to different sampling rates (i.e. $l_{i,j_1} \neq l_{i,j_2}$) or both. The goal is to find a subset ${j_1,...,j_k} \subset [1..m]$ of sensors that that are (1) predictive of the labels and (2) have low redundancy.

The distance between two time-series $\bm{x_{i_1,j}}$ and $\bm{x_{i_2,j}}$ can be calculated using Dynamic Time Warping (DTW) \cite{berndt1994using, keogh2005exact}. For example, given two time series:
\begin{align*}
\centering
S &= s_1, s_2, \cdots, s_i, \cdots, s_n\\
T &= t_1, t_2, \cdots, t_j, \cdots, t_m,
\end{align*}
the sequences $S$ and $T$ can be arranged to form a $n$-by-$m$ grid, where each grid point, $(i,j)$, corresponds to an alignment between elements $s_i$ and $t_j$. A warping path, $P$, maps the elements of $S$ and $T$, such that the ``distance'' between them are minimized:
\begin{equation*}
P = p_1, p_2, \cdots, p_K, \quad max(n,m) \leq K < n+m,
\end{equation*}
where $K$ is the length of wrap path and the $k^{th}$ warp path is:
\begin{equation*}
    p_k = (i,j),
\end{equation*}
where $i$ is an index from time series $S$ and $j$ is an index from time series $T$. The warp path start from $p_1=(1,1)$ and end at $p_K=(n,m)$. The index $i$ and $j$ have to be monotonically increasing in the warp path.

If we define a distance between two elements, such as:
\begin{equation*}
    \delta(i,j) = |s_i - t_j|.
\end{equation*}
The DTW distance of two time series is the optimal warp path with minimum distance:
\begin{equation*}
    DTW(S,T) = \min_{P}[\sum_{k=1}^{q}\delta({p}_k)],
\end{equation*}
where $q$ is the length of optimal warping path.

\begin{table}[]
    \centering%
    \begin{tabular}{|l|l|l|}%
    \hline%
        Symbol & Dimension & Meaning \\ \hline
        $\bm{X}_i$ &                    & MTS \\ \hline
        $\bm{x}_{i,j}$ &  $\mathcal{R}^{m \times l_{i,j}}$ & time-series \\ \hline
        $y_i$ & $\mathcal{R}$ & data label \\ \hline
        $\bm{y}$ & $\mathcal{R}^{n}$ & label vector \\ \hline
        $\bm{M}_j$ & $\mathcal{R}^{n\times n}$ & distance matrix of $j$-th row \\ \hline
        $\bm{W}_j$ & $\mathcal{R}^{n\times n}$ & similarity graph of $j$-th row \\ \hline
        $\bm{D}$ & $\mathcal{R}^{n\times n}$ & degree matrix \\ \hline
        $\bm{v}$ & $\mathcal{R}^{n} $ & power iteration embedding \\ \hline
        $\bm{L}$ & $\mathcal{R}^{n} $ & graph Laplacian \\ \hline
        $\bm{G}_I$ & $\mathcal{R}^{n\times n}$ & redundancy constraint matrix \\ \hline
        $\bm{Q}$ & $\mathcal{R}^{n\times n}$ & redundancy constraint matrix \\ \hline
        $\bm{H}$ & $\mathcal{R}^{n^2\times m}$ & flattened similarity matrix \\ \hline
    \end{tabular}
    \caption{Math notations.}
    \label{tab:my_label}
\end{table}

\section{Algorithm}
\label{sec:algorithm}
\subsection{Build Similarity Graph for Sensors}
As mentioned, we build a similarity graph for each sensor of $\bm{X}_i$. There exists many graph construction algorithms in machine learning research area, for example, $k$-nearest neighbor graph, sparse graph~\cite{liu2010large} and etc. In our work, we choose \textit{$k$-nearest neighbor graph} as it can preserve the local geometry structure of original data's distribution. The first step (1) is to calculate the distance between two time series. We assume our time series data are real values and use the Dynamic Time Warping~\cite{berndt1994using}~\cite{keogh2005exact} metric. (Note: Our work can easily be extended to non-real value data, as long as the user can define a distance metric. For example, for string representation, the Levenshtein distance (or Edit distance) can be used.) (2) once we have the distance matrix $\bm{M}_j$ for $j$-th row of $\bm{X}$, we generate the $k$-nearest neighbor graph $\bm{W}_j$ which is directed graph with edge weight equal to ``1.0". The reason we reset the distance is for the purpose of normalization since the DTW distance bound of different sensors are quite uncertain. After that, we transfer the directed graph into an undirected one by:
\begin{equation}
    \bm{W}_j = 0.5*(\bm{W}_j + \bm{W}^T_j),
\end{equation}
then we obtain a similarity graph with symmetric adjacency matrix. Following the spectral clustering work~\cite{shi2000normalized}, we set the diagonal of $\bm{W}_j$ to zero.

To summary, the output of this graph construction step is a set of matrices: $\left\{\bm{W}_1, \bm{W}_2, \cdots, \bm{W}_m, \bm{W}_y\right\}$, where $\bm{W}_j \in \mathcal{R}^{n\times n}$.

\subsection{The Structure Representation Vector of a Graph}
In this section, we introduce a spectral embedding vector which encodes the cluster  structure information of similarity graphs we constructed in last section. We believe spectral graph embedding is a right approach as the performance shown in spectral clustering~\cite{shi2000normalized} and unsupervised feature selection~\cite{cai2010unsupervised}. Among many possible choices such as Power Iteration Embedding (PIE)~\cite{lin2010power}, Top-$k$ Spectral Clustering Embedding~\cite{shi2000normalized} and Heat Kernel Embedding~\cite{peng2015partitioning}, we choose PIE as our spectral embedding vector by its ``1-D" dimensional vector property and its simplicity of calculation. The description of PIE is as follows.

The power iteration embedding embedding vector is an early stopping approximation of the largest eigenvector of normalized affinity matrix $\bm{W}$ which equals to $\bm{D}^{-1}\bm{W}$, where $\bm{D}$ is the degree matrix of $\bm{W}$. The power iteration embedding $\bm{v}^t$ can be calculated as:
\begin{equation}
    \bm{v}^t = c\bm{D}^{-1}\bm{W}\bm{v}^{t-1},
\end{equation}
where $c$ is a normalizing constant to limit the value of $\bm{v}$, and is set as $c=\|\bm{D}^{-1}\bm{W}\bm{v}^{t-1}\|_1^{-1}$. mbedding approximates the cluster structure of graph by using an one dimensional vector. The detailed algorithm is described in Alg.~\eqref{alg:PIE}.

\begin{algorithm}[h!]
\SetAlgoLined
\KwIn{matrix $\bm{W}\in\mathcal{R}^{n\times n}$.}
\KwOut{power iteration embedding $\bm{v}^t\in\mathcal{R}^{n\times 1}$.}
Apply positive random normalization: $\bm{W} \gets \bm{D}^{-1}\bm{W}$\;
Initialize $\bm{v}^{0}\in\mathcal{R}^{n\times 1}$\;
Repeat\;
  $\bm{v}^{t+1} \gets \frac{\bm{W}\bm{v}^t}{\|\bm{W}\bm{v}^t\|_1}$\;
  $\delta^{t+1} \gets |\bm{v}^{t+1} - \bm{v}^{t}|$\;
  $t \gets t+1$\;
Until $\|\delta^t - \delta^{t+1}\|_{max} \simeq 0$\;
\Return{$\bm{v}^t$}
\caption{PIE($\bm{W}$)~\cite{lin2010power}}
\label{alg:PIE}
\end{algorithm}

\subsection{Sensor Ranking}
\label{sec:feature_ranking}
Our first task is to rank the importance of sensors according to the labels. We introduce a filter method named as ``PIE-rank'' which includes two steps: (1) calculates the PIE embedding vector of each sensor's similarity graph, and (2) evaluates the Normalized Mutual Information (NMI) score between the PIE embedding vector and ground truth label vector $\bm{y}$. The NMI score is calculated as follows.
\begin{equation}
    r(\bm{v}^t) = \text{score}(\bm{v}^t, \bm{y}) = \text{NMI}(\bm{v}^t, \bm{y}) = \frac{I(\bm{v}^t, \bm{y})}{\sqrt{H(\bm{v}^t) H(\bm{y})}},
    \label{eq:nmi}
\end{equation}

where $I(\cdot,\cdot)$ denotes the mutual information , and $H(\cdot)$ denotes the entropy. When calculate the mutual information between $\bm{v}^t$ and $\bm{y}$, $k$-means clustering is used to put the real values of $\bm{v}^t$ into different bins. The number of bins equals to the number of different labels in $\bm{y}$. Higher ranking score means the spectral embedding vector is more close to the label's distribution. 

The sensor ranking algorithm (PIE-rank) can be summarized by:
\begin{algorithm}[h!]
\SetAlgoLined
\KwIn{MTS and labels: $\{(\bm{X}_i, y_i)\}$.}
\KwOut{Ranked scores $\bm{r}$.}
\For{$i \leftarrow 1$ \KwTo $m$} {
    Build the distance graph $\bm{M}_i$\;
    Generate the $k$-nearest neighbor graph $\bm{W}_i$\;
    $\bm{W}_i$ = $0.5*(\bm{W}_i + \bm{W}^T_i$)\;
    $\bm{v}^t$ = PIE($\bm{W}_i$)\;
    $\bm{r}(i)$ = scores($\bm{v}^t$, $y$)\;
}
Sort($\bm{r}$)\;
\Return{$\bm{r}$}
\caption{PIE-rank}
\label{alg:PIE-fs}
\end{algorithm}%

\subsection{Sensor Subset Selection}
The sensor ranking algorithm introduced in previous Section~\ref{sec:feature_ranking} has very straightforward result and characterized by its simplicity and efficiency. However, the redundancy existing in the top selected sensors~\cite{yu2004efficient} are not handled. The redundancy~\cite{peng2005feature} means there exists several sensors which are highly correlated to each other. To minimize the redundancy, we present the sensor subset selection algorithm ``PIE-SS'' (``SS'' means ``Subset Selection'') in this section.

\subsection{Object function} 
Our learning goal is to select a subset of sensors with minimum redundancy. Our intuition is to search a sparse linear combination of each sensor's similarity graph, and let the combined new graph can approximate the distribution of labels (represented by a label graph) as much as possible.At the same time, the selected subset sensors should have minimum redundancy w.r.t the NMI value among their PIE embedding vectors.

One challenge here is that the redundancy constraint is in the spectral space while the linear approximation of similarity graph is in original data space. It is possible that two different sensor graphs will have same spectral embedding vector. For example, for two graphs have the same cluster structure but different edge connection patterns in each cluster, they spectral embedding vectors will be very similar to each other. 

Base on the above learning goal and observations, we propose following object function to calculate the optimal sensor subset with minimum redundancy:\\
\begin{multline}
\min_{\alpha}\frac{1}{2}\|\bm{W}_y - \sum_{i=1}^{m}\alpha_i\bm{W}_i\|_F^2 + \lambda\sum_{i=1}^{m}|\alpha_i| \\ + \beta\sum_{i,j}\alpha_i\alpha_j I(\bm{v}^t_i, \bm{v}^t_j), \quad \text{s.t. } \alpha_i \geq 0. \label{eq:obj}
\end{multline}
where $\lambda \geq 0, \beta \geq 0$ are Lasso penalty and redundancy control parameters. The first term represents the approximation of label graph by the linear combination of sensor graphs. The second term put the sparse constraint to the coefficients of linear combination. The third term add redundancy constraints to the selection of sensor graph in final subset. 

The objective function can be rewrote in matrix form as:
\begin{equation}
\min_{\alpha}\frac{1}{2}\|\bm{W}_y - \bm{W}\bm{\alpha}\|_F^2 + \lambda\|\bm{\alpha}\|_1 + \beta\bm{\alpha}^T\bm{G}_I\bm{\alpha}. 
\label{eqn:obj_matrix}
\end{equation}
The proposed object function is non-convex by nature as the matrix $\bm{G}_I$ is not  a positive semi-definite one (even it is close to)~\cite{jakobsen2014mutual}. Moreover, the diagonal elements of $\bm{G}_I$ also let the object function to penalty the ``self-redundancy" $\bm{G}_I(i,i)$ and create selection bias in favor of sensor graph whose embedding vector has lower entropy as pointed out in ~\cite{nguyen2014effective}. Following the treatment in ~\cite{nguyen2014effective}, we introduce the matrix $\bm{Q}$ and hyper-parameter $\lambda$ to equation~\eqref{eqn:obj_matrix}. The matrix $\bm{Q}$ is defined as:  
\begin{equation}
 \bm{Q}_{ij} =
  \begin{cases}
    \bm{I}(\bm{v}_i^t, \bm{y})       & \quad \text{if } i = j\\
    \frac{1}{2}(\bm{I}(\bm{v}_i^t;\bm{y}|\bm{v}_j^t) + \bm{I}(\bm{v}_j^t;\bm{y}|\bm{v}_i^t))  & \quad \text{if } i \neq j.
  \end{cases}
\end{equation}
The improved object function then becomes:
\begin{align}
\mathcal{J}(\bm{\alpha})=
\frac{1}{2}\|\bm{W}_y - \bm{W}\bm{\alpha}\|_F^2 + \lambda\|\bm{\alpha}\|_1 + \beta\bm{\alpha}^T(\bm{Q}+\gamma\mathcal{I})\bm{\alpha},\nonumber \\ 
\label{eqn:obj_matrix_Q}
\end{align}
where $\mathcal{I}$ is the identity matrix who has the same size as $\bm{Q}$.

\subsection{Calculation of $\bm{Q}$ for Large Data} 
\label{sec:build_q}
The calculation of mutual information is time consuming, and it is quite difficult to finish the calculation of matrix $\bm{Q}$ in a short time. To solve this scalability issue, we apply the Nystrom matrix approximation following the strategy in~\cite{nguyen2014effective}. 
\begin{equation}
    \tilde{\bm{Q}} = \begin{bmatrix}
    A & B \\
    B^T & B^TA^{-1}B
    \end{bmatrix}
\end{equation}
\noindent\textbf{$\tilde{\bm{Q}}$} and Positive Semidefinite(PSD) Matrix. Our first step is to pre-processing the matrix $\tilde{\bm{Q}}$ by adjusting the parameter $\gamma$. The goal is to improve $\tilde{\bm{Q}}$ to $\bm{\hat{Q}}$ which is a PSD matrix.\\

\subsubsection{Coordinate Descent Solver}
To solve Equation~\eqref{eqn:obj_matrix_Q}, we apply the Coordinate Descent algorithm~\cite{friedman2010regularization}~\cite{wright2015coordinate} by its scalability. We flatten each matrix $\bm{W} \in \mathcal{R}^{n\times n}$ into a vector representation $\bm{h}\in\mathcal{R}^{n^2\times1}$. This will help to better understand steps of calculating derivatives. Now Equation \eqref{eqn:obj_matrix_Q} is changed to:
\begin{equation}
    \mathcal{J}(\bm{\alpha})=
\frac{1}{2}\|\bm{h}_y - \bm{H}\bm{\alpha}\|_2^2 + \lambda\|\bm{\alpha}\|_1 + \beta\bm{\alpha}^T\hat{\bm{Q}}\bm{\alpha},\nonumber \\ 
\end{equation}
where $\bm{H}\in \mathcal{R}^{n^2\times m}$ and each column of $\bm{H}$, $\bm{H}_j$, is the flat version of $\bm{W}_j$, $\bm{h}_y \in \mathcal{R}^{n^2\times1}$, $\bm{\alpha}\in\mathcal{R}^{m\times1}$ and $\hat{\bm{Q}}\in \mathcal{R}^{m\times m}$. \\

The object function $\mathcal{J}(\bm{\alpha})$ can be split into two parts as:
\begin{align}
    \mathcal{J}(\bm{\alpha}) &= f(\bm{\alpha}) + g(\bm{\alpha}) \\
    f(\bm{\alpha}) & = \frac{1}{2}\|\bm{h}_y - \bm{H}\bm{\alpha}\|_2^2 + \beta\bm{\alpha}^T\hat{\bm{Q}}\bm{\alpha} \\
g(\bm{\alpha}) &= \lambda\|\bm{\alpha}\|_1.
\end{align}

We separate each term into $\alpha_k$ part and $\alpha_{-k}$ part. The notation `${-k}$' means the set does not include $k$-th item.

\begin{align*}
    f(\bm{\alpha}) &= \frac{1}{2}\|\bm{h}_y - \bm{H_{\cdot,-k}}\bm{\alpha_{-k}} - \bm{H}_{\cdot,k}\alpha_k\|_2^2 \\ 
    &+ \beta\bm{\alpha}_{-k}\hat{\bm{Q}}_{-k,-k}\bm{\alpha}_{-k} \\ &+ \beta \alpha_k(\sum_{j}\hat{\bm{Q}}_{kj}\bm{\alpha}_j) + \beta(\sum_{i}\bm{\alpha}_i\hat{\bm{Q}}_{ik})\alpha_k,\nonumber 
\end{align*}
The derivative of $f$ over one coordinate $\alpha_k$ is:
\begin{align}
\label{equ:obj_deriv}
    \frac{\partial f}{\partial\alpha_k} &= -\bm{H}_k^T(\bm{h}_y - \bm{H}_{\cdot,-k}\bm{\alpha}_{-k} - \bm{H}_{\cdot,k}\bm{\alpha}_k) \nonumber\\
    &\quad +\beta(\sum_{j}\hat{\bm{Q}}_{kj}\bm{\alpha}_j) + \beta(\sum_{i}\bm{\alpha}_i\hat{\bm{Q}}_{ik}), \nonumber\\
    &= -\bm{H}_k^T(\bm{h}_y-\bm{H}\bm{\alpha}) \nonumber\\
    &\quad + \beta\hat{\bm{Q}}_{k,\cdot}\bm{\alpha} + \beta\bm{\alpha}^T\hat{\bm{Q}}_{\cdot,k}. \nonumber \\
    &= \bm{H}_k^T\bm{H}\bm{\alpha} - \bm{H}_k^T\bm{h}_y + 2\beta\bm{\alpha}^T\hat{\bm{Q}}_{\cdot,k}. \nonumber \\
\end{align}

\subsubsection{Proximal Operator} The componentwise proximal operator of $g(\alpha_i)$ which is in $\ell_1$-norm is:
\begin{equation}
    \textrm{Prox}_{\lambda\|\cdot\|_1}(\alpha_i) = \arg\min_{\hat{\alpha}}\frac{1}{2}\|\hat{\alpha}_i-\alpha_i\|_2^2+\lambda\|\hat{\alpha}\|_1,
\end{equation}
and the solution is:
\begin{equation}
    \hat{\alpha}_i^{lasso} = \textrm{sign}(\hat{\alpha}_i)\max(|\hat{\alpha}_i|-\lambda, 0)
\end{equation}

\subsubsection{Scalability} For large dataset, which means the MTS has many sensors in our case, the running time of regular coordinate descent algorithm is kind of challenge. To improve the scalability, we take the advantage of multi-core architecture of modern computer system and the power of statistic optimization. To be specific, we apply the asynchronous stochastic coordinate descent algorithm proposed in~\cite{liu2015asynchronous}.

\subsubsection{PIE-SS} With the solution of object function~\eqref{eq:obj}, we present the ``PIE-SS'' as in~\eqref{alg:PIE-ss}. 
\begin{algorithm}[h]
\SetAlgoLined
\KwIn{MTS and labels: $\{(\bm{X}_i, y_i)\}$.}
\KwOut{Subset of features.}
\For{$i \leftarrow 1$ \KwTo $m$} {
    build the distance graph $\bm{M}_i$\;
    generate the $k$-nearest neighbor graph $\bm{W}_i$\;
    $\bm{W}_i$ = $0.5*(\bm{W}_i + \bm{W}^T_i$)\;
    $\bm{v}^t$ = PIE($\bm{W}_i$)\;
    Solve the object function~\eqref{eq:obj} and obtain results $\bm{\alpha}$\;
    Select features that their coefficients are larger than zero: $\bm{\alpha}_i > 0$\;
}
\Return{Selected features.}
\caption{PIE-SS}
\label{alg:PIE-ss}
\end{algorithm}%

\section{Connection to Two-Stage Multiple Kernel Learning}
\label{sec:discussion}
Our proposed feature subset selection algorithm has strong connection to two stage multiple kernel learning~\cite{kumar2012binary}~\cite{cortes2010two}~\cite{kandola2002optimizing} and kernel-target alignment~\cite{cristianini2002kernel} in theory. The target alignment problem is defined as follows.
\begin{equation}
    \max_{\mu \geq 0} \frac{\langle \sum_{l=1}^p \mu_l\bm{K}_l, \bm{K}^{(t)} \rangle}{\|\sum_{l=1}^{p}\mu_l \bm{K}_l\|_F}, \textrm{s.t.} \quad \|\mu\|_2=1, 
\end{equation}
where $A$ is the Gram matrix of kernel $A$ on the training samples, $\langle A,B \rangle = tr(AB^T)$ and $\|A\|_F^2=tr(AA^T)$. The $\bm{K}^{(t)}$ is the target kernel, and $\bm{K}_l$ is predefined kernel from a set which has size equals to $p$.

The problem of feature subset algorithm in this works is similar to two-stage kernel learning problem. Both of them learn a non-negative linear combination of base kernels that maximizes the alignment with one target kernel. The similarity graph $\bm{W}_y$ in object function~\eqref{eq:obj} can be treated as a target kernel $\bm{K}^{(t)}(\bm{X}_i, \bm{X}_j)=y_iy_j$ in binary classification case. For multiple label classification cases, it is natural to use ``one-vs-all'' strategy to redefine the problem.

The most obvious difference is that two-stage kernel learning problem define base kernels at instance side while our feature subset selection algorithm define base kernels at feature side. The two-stage kernel learning problem tries to search the optimal linear combination of a set of predefined kernel (size is specified by the user) on instances, and hope the combined kernels can show a data distribution as similar as target kernels which represent the distribution of labels. Contrarily, the feature subset selection problem in this work defines kernels (or similarity graphs in current setting) for each single feature, and the number of predefined kernels is equal to the number of features. The k-nearest neighbor graphs used in this work can be extended to other popular kernels such as Gaussian kernel, Epanechnikov Kernel and so on. 

Another difference is that the feature subset selection problem does not include the learning of classifier/regressor which is the second stage of two-stage kernel learning problem. Since the focus of this work is feature selection.

\section{Experiments}
\label{sec:experiments}
\subsection{Datasets}
Our experiment results includes seven public available MTS datasets. One of them, named as ``BasicMotions'', is used for illustration purpose only and be excluded from comparisons. The remained six datasets are used for evaluation. Two for them, ``DuckDuckGeese'' and ``PEMS-SF'', are from the UEA \& UCR time series classification repository~\cite{bagnall16bakeoff}. The ``EGG'' (large) dataset is from UCI~\cite{Dua:2019}. The ``CMU-MOCAP-S16'' and ``KickvsPunch'' is from ~\cite{mustafa}. The ``HumanGait'' dataset is from~\cite{tanawongsuwan2003performance}. A summary of them is presented in Table~\eqref{tab:data_mts}.
\begin{table}[h]
    \centering
    \resizebox{\columnwidth}{!}{%
    \begin{tabular}{|*{6}{c|}}
    \toprule
        Name & Train & Test & \#Sensors & Length  & Class \\ \midrule
        BasicMotions & 40 & N/A & 6 & 100 & 4 \\
        DuckDuckGeese & 60 & 40 & 1345 & 270 & 5 \\
        EEG (UCI) & 468 & 480 & 64 & 256 & 2 \\
        CMU-MOCAP-S16 & 29 & 29 & 62 & 127-580 & 2 \\
        KickvsPunch & 16 & 10 & 62 & 274-841 & 2 \\
        HumanGait & 270 & 270 & 66 & 133 & 15 \\
        PEMS-SF & 267 & 173 & 963 & 114 & 7 \\ \bottomrule
    \end{tabular}}
    \caption{Summary of selected MTS datasets. The ``BasicMotions" dataset is used for illustration in Fig.~\eqref{fig:MTS_PIE}. For ``CMU-MOCAP-S16'' and ``KickvsPunch'', the length of different sensors are not the same.}
    \label{tab:data_mts}
\end{table}
\subsection{Baseline Algorithms} To evaluate the performance of our proposed algorithms, we select two famous algorithms CLeVer~\cite{yoon2005feature} and Corona~\cite{yang2005supervised} as our baseline. We also aware that there exists one latest work CSFS~\cite{han2013feature} which has the same idea as Corona but use the Mutual Information instead of Correlation to vectorize each MTS. However, the calculation of Mutual Information for MTS with large number of sensors is a problem as we discussed in Section~\ref{sec:build_q}, let alone the CSFS need to repeat the calculation for each data sample. By this reason, we drop our comparison with it.

\begin{itemize}
    \item CLeVer: the CLeVer algorithm uses the loadings of common principal component analysis to measure the importance of each sensor. First, a correlation coefficient matrix is built among different sensors for each MTS segment. Secondly, principal components of each coefficient matrix are calculated. Thirdly, all these principal components are aggregated together and the descriptive common principal components are calculated. Finally, the $\ell^2-norm$ of loading vectors are used to rank the importance of each sensor.
    \item Corona: the Corona algorithm uses the flattened correlation coefficient matrix as feature vector and recursive feature elimination~\cite{guyon2002gene} to rank sensors. The coefficients of trained support vector machine are used to indicate the importance of each sensors during the iteration. 
\end{itemize}

\begin{figure*}[ht!]
    \centering
    \begin{minipage}[t]{0.32\linewidth}
    \centering
    \includegraphics[width=\textwidth]{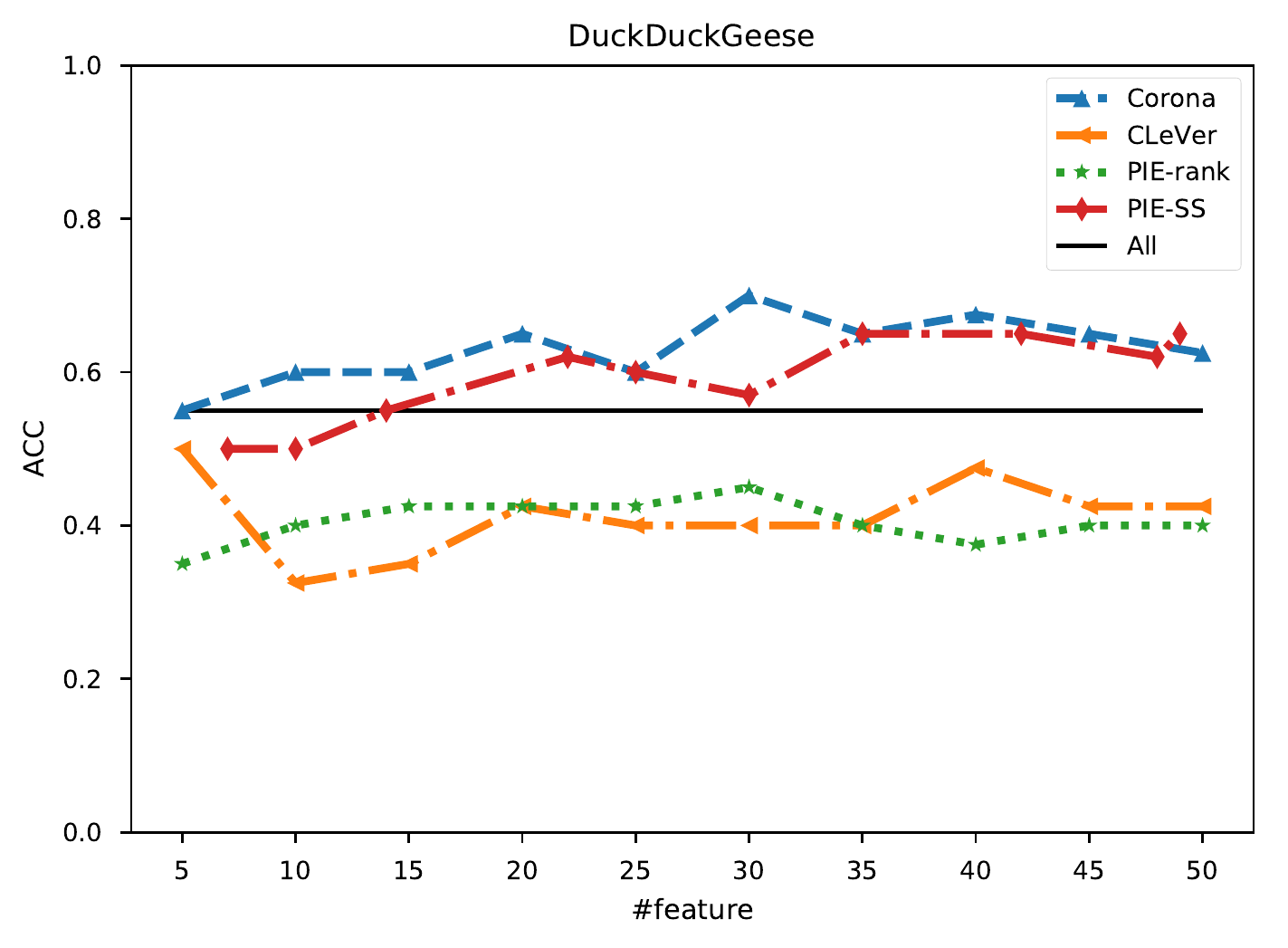}
    \end{minipage}
      \begin{minipage}[t]{0.32\linewidth}
    \centering
    \includegraphics[width=\textwidth]{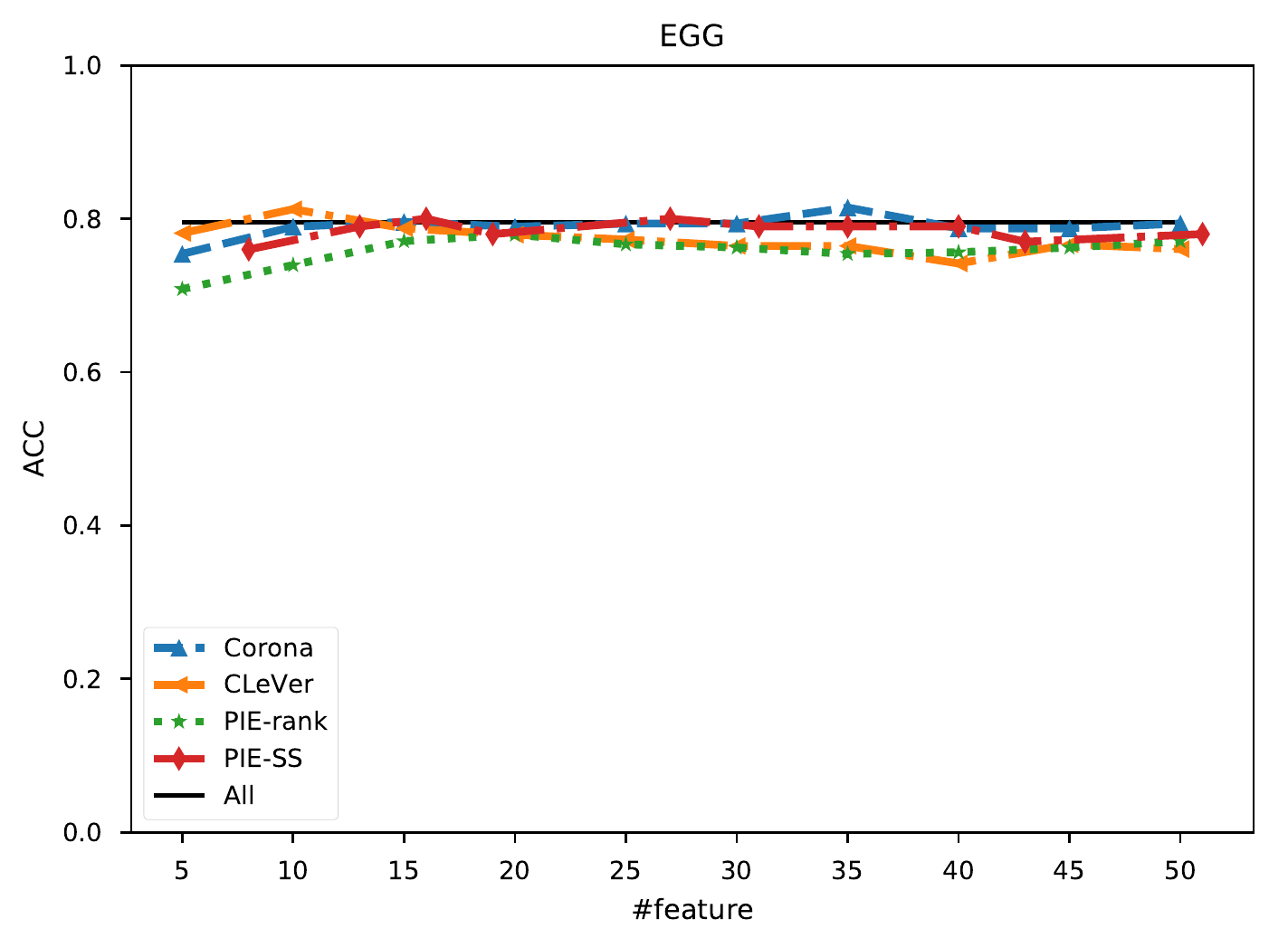}
    \end{minipage}
      \begin{minipage}[t]{0.32\linewidth}
    \centering
    \includegraphics[width=\textwidth]{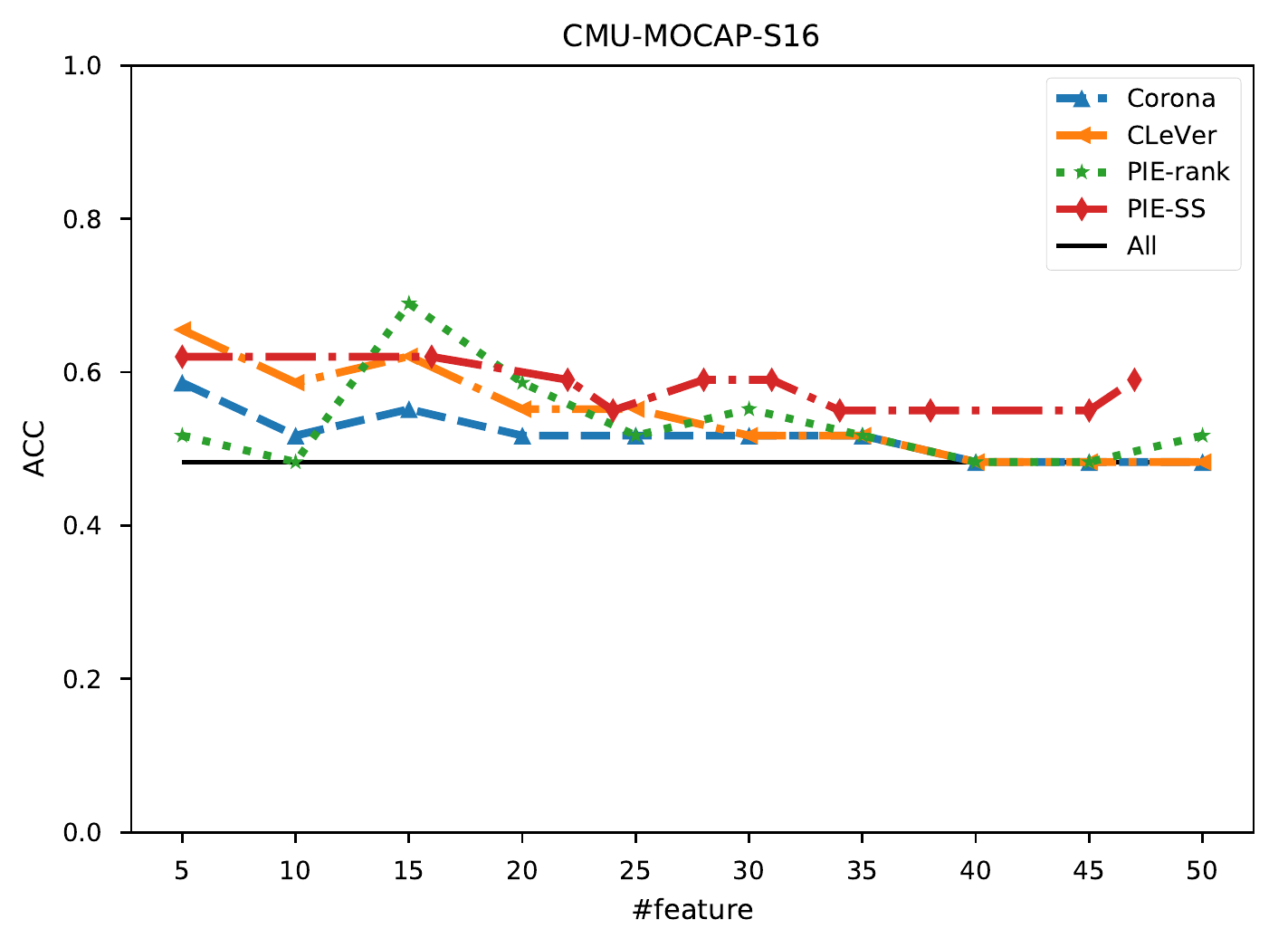}
    \end{minipage}\\%
      \begin{minipage}[t]{0.32\linewidth}
    \centering
    \includegraphics[width=\textwidth]{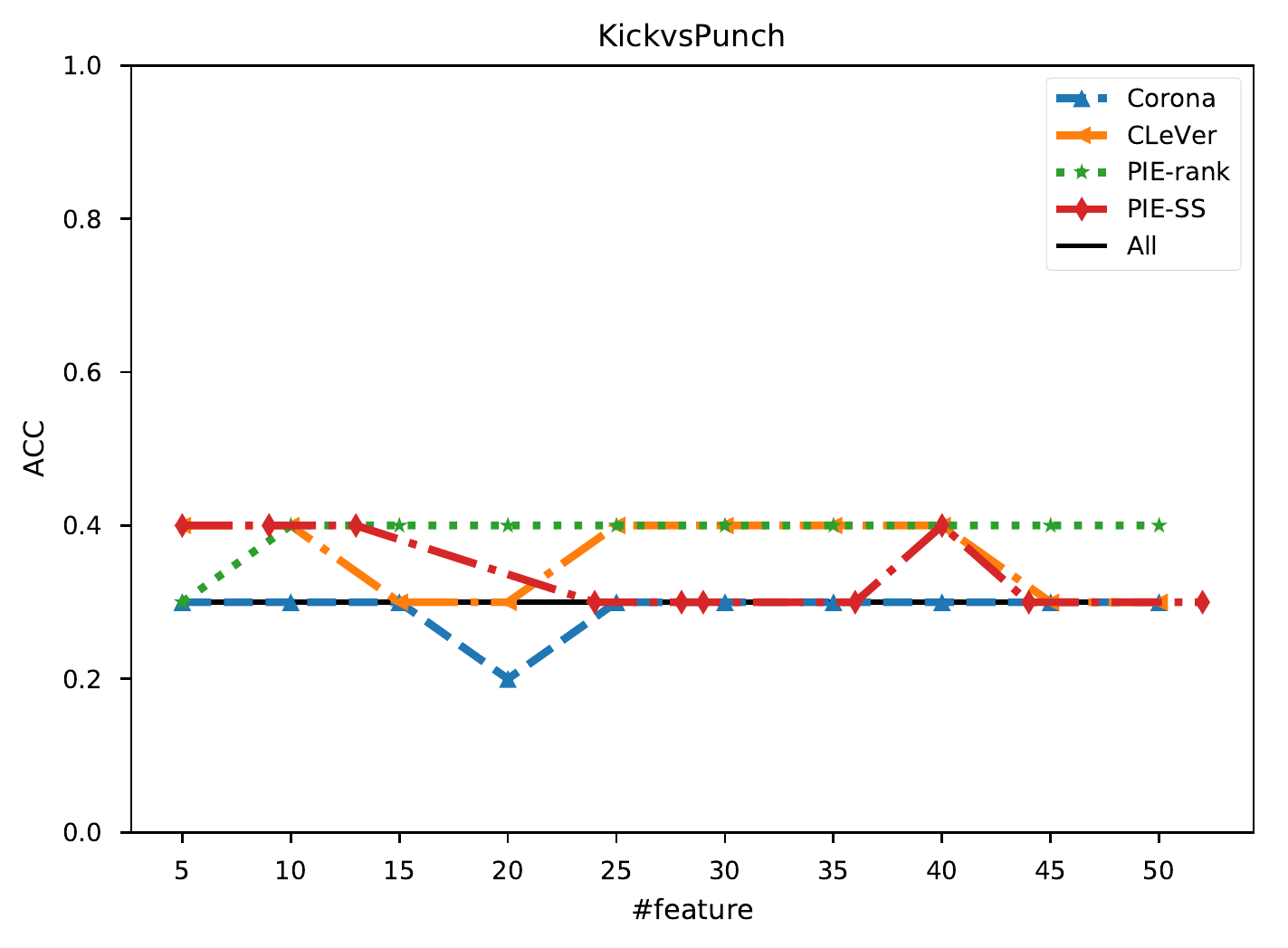}
    \end{minipage}
      \begin{minipage}[t]{0.32\linewidth}
    \centering
    \includegraphics[width=\textwidth]{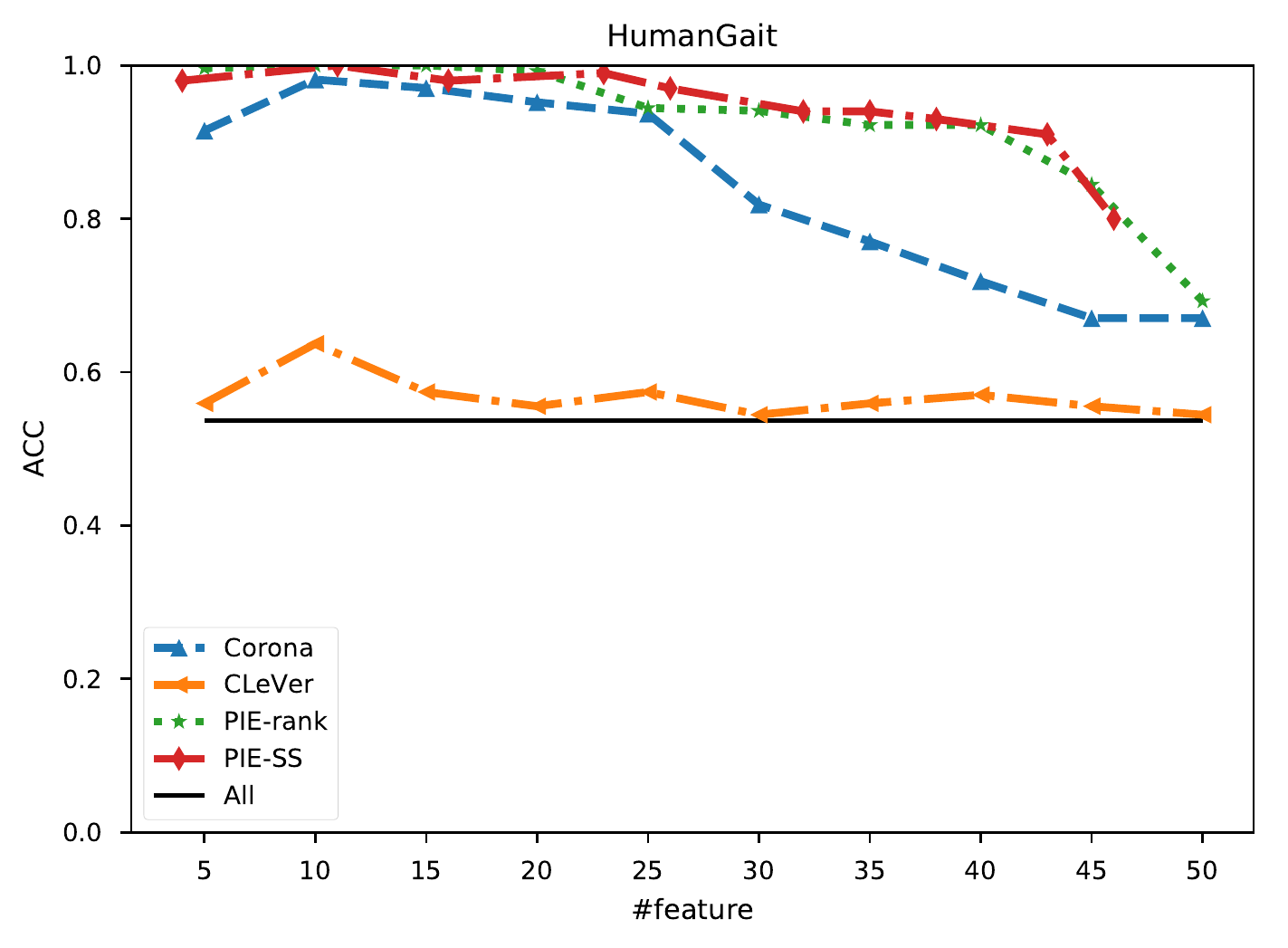}
    \end{minipage}
    \begin{minipage}[t]{0.32\linewidth}
    \centering
    \includegraphics[width=\textwidth]{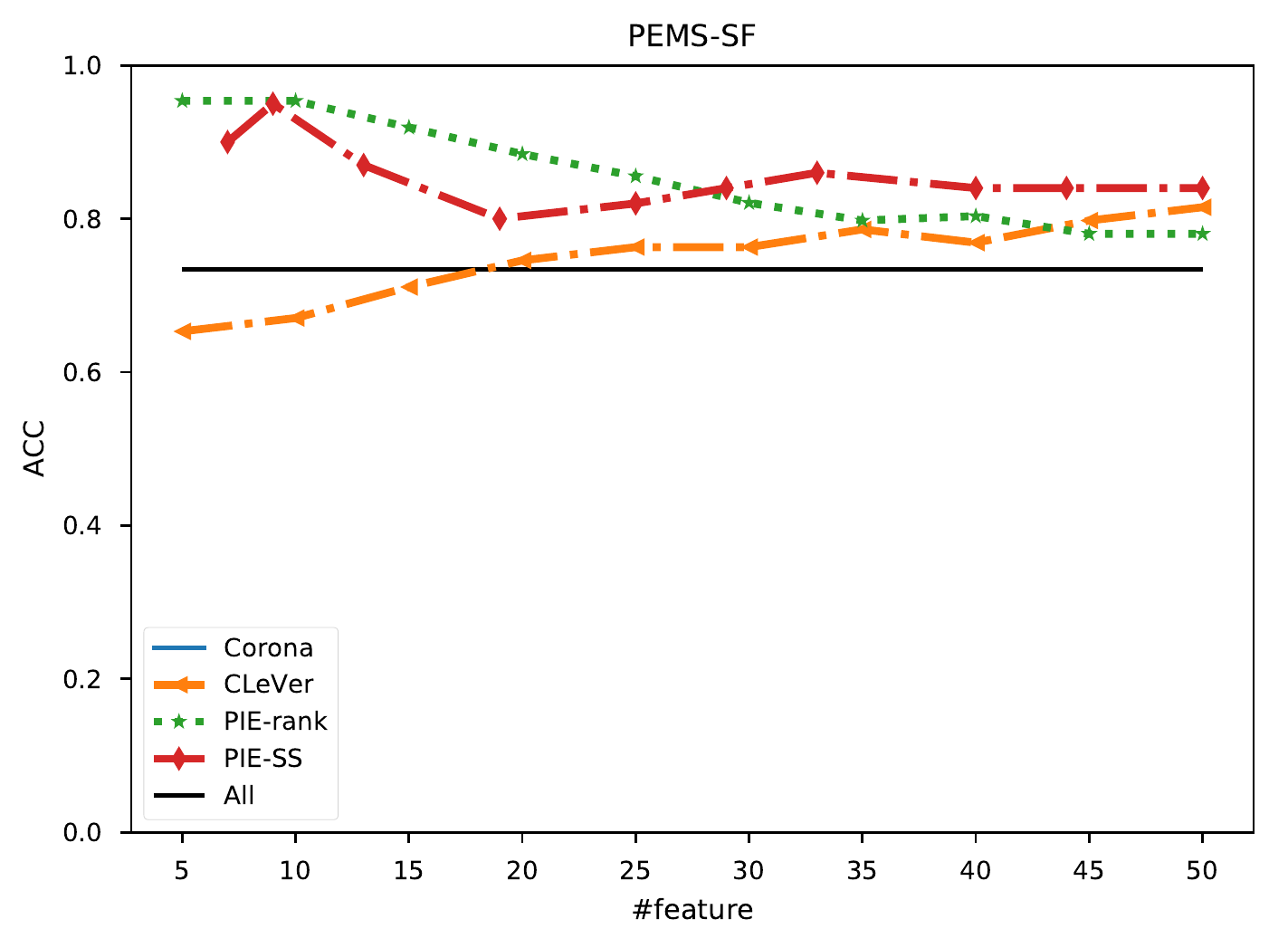}
     \end{minipage}
    \caption{Classification results of different algorithms on selected MTS data. The ``unweighted'' version results of ``PIE-SS" is reported here. Specially, for the PEMS-SF dataset, the results of ``Corona'' algorithm is missing as the time of calculating it is more than 24 hours (with our implementation).}
    \label{fig:fea_fs_mts}
\end{figure*}%

\begin{table*}[ht]
    \centering
    \resizebox{\linewidth}{!}{%
    \begin{tabular}{|*{12}{c|}}
    \toprule
        Name/ACC  & \textit{dists} & Size 1 & Size 2 & Size 3 & Size 4 & Size 5 & Size 6 & Size 7 & Size 8 & Size 9 & Size 10\\ \midrule
CMU-MOCAP-S16	 & W & \textbf{0.62}(5)	& \textbf{0.62}(16) & 0.55(22)	& 0.55 (24) & 0.55 (28) &	0.55(31) & 0.59(34) & 0.55(38)& 0.55(45) & 0.59(47)\\
CMU-MOCAP-S16	& UW & \textbf{0.62}(5) & \textbf{0.62}(16) & 0.59(22)	& 0.55 (24) & 0.59 (28) &	0.59(31) & 0.55(34) & 0.55(38)& 0.55(45) & 0.59(47)\\ \hline
DuckDuckGeese & W & 0.53(8) & 0.50(10) &0.55(14) & 0.57(22) & 0.53(25) & 0.57(30) & 0.62(35) & 0.62(42) & 0.60(48) & \textbf{0.65}(49)  \\
DuckDuckGeese & UW & 0.50(8) & 0.50(10) &0.55(14) & 0.62(22) & 0.60(25) & 0.57(30) & \textbf{0.65}(35) & \textbf{0.65}(42) & 0.62(48) & 0.65(49)  \\ \hline
EGG	& W &0.77(8) & 0.80(13) &0.82(16) & 0.81(19) & 0.81(27) & 0.80(31) & 0.80(35) & 0.79(40) & \textbf{0.83}(43) & 0.79(51)  \\
EGG	& UW &0.76(8) & 0.79(13) & \textbf{0.80}(16) & 0.78(19) & \textbf{0.80}(27) & 0.79(31) & 0.79(35) & 0.79(40) & 0.77(43) & 0.79(51)  \\ \hline
KickvsPunch	& W & \textbf{0.60}(5) & 0.40(9) &0.50(13) & 0.50(24) & 0.3(28) & 0.40(29) & 0.30(36) & 0.4(40) & 0.40(44) & 0.40(52) \\
KickvsPunch	& UW & \textbf{0.40}(5) & \textbf{0.40}(9) & \textbf{0.40}(13) & 0.30(24) & 0.3(28) & 0.30(29) & 0.30(36) & 0.4(40) & 0.30(44) & 0.30(52) \\ \hline
HumanGait & W &  0.98(4) & \textbf{1.00}(11) & \textbf{1.00}(16) & \textbf{1.00}(23) & \textbf{1.00}(26) & \textbf{1.00}(32) & \textbf{1.00}(35) & \textbf{1.00}(38) & 0.99(43) & 0.97(46) \\
HumanGait & UW &  0.98(4) & \textbf{1.00}(11) & 0.98(16) & 0.99(23) & 0.97(26) & 0.94(32) & 0.94(35) & 0.93(38) & 0.91(43) & 0.80(46) \\ \hline
PEMS-SF	& W & \textbf{0.95}(7) & \textbf{0.95}(9) & {0.89}(13) & {0.88}(19) & {0.87}(25) & {0.87}(29) & {0.88}(33) & {0.88}(40) & 0.87(44) & 0.87(50) \\
PEMS-SF	& UW & 0.90(7) & \textbf{0.95}(9) & {0.87}(13) & {0.80}(19) & {0.82}(25) & {0.84}(29) & {0.86}(33) & {0.84}(40) & 0.84(44) & 0.84(50) \\ \bottomrule
    \end{tabular}}%
    \caption{Classification performance by PIE-SS with different size of selected sensors. The size of selected sensors are put inside the parenthesis. Results include two different settings of aggregating distance matrix. They are marked as ``UW'' for `$\textit{dist}_{unweighted}$'' and ``W'' for `$\textit{dist}_{unweighted}$''. The integer value in the parenthesis after each accuracy value means the size of selected sensors by the PIE-SS algorithm. The highest value of each row is marked as bold.}
\end{table*}

\subsection{Evaluation Classifier and Metrics} We use one nearest neighbor (NN-1) classifier as our benchmark algorithm following the UEA multivariate time series website~\cite{bagnall2018uea}. Moreover, to measure the performance of PIE-SS, we use two different settings to aggregate the selected features (sensors). One is a unweighted version which can be found in most feature selection research works, another is a weighted version since we learn a linear combination though Equation~\eqref{eq:obj}. The details are as follows.

\begin{enumerate}
    \item Unweighted aggregation of sensors:
    \begin{equation*}
    \textit{dist}_{unweighted} =\sum_{i=1}^{k}\bm{W}_i.
    \end{equation*}
    \item Weighted aggregation of sensors:
     \begin{equation*} 
     \textit{dist}_{weighted} = \sum_{i=1}^{k}\alpha_i\bm{W}_i,
     \end{equation*}
\end{enumerate}
where $\textit{dist}_{*}$ is the aggregated distance matrix and will be used for nearest neighbor classifier, $k$ is the number of selected sensors.

For the evaluation metric, we use Accuracy(ACC).

\subsection{Parameter Settings}
We list the configuration of parameters (if any) for each algorithm as below.
\begin{itemize}
    \item CLeVer: we set the $\delta = 0.7$ as in Alg. 1 in~\cite{yoon2005feature}. 
    \item PIE-rank: when calculating the $k$-nearest neighbor graph of each sensor, we set $k$ equal to 10. 
    \item PIE-SS: the $\beta$ is set to 1.0. The $\lambda$ parameter is used to control the sparsity of results. 
\end{itemize}

\subsection{Discussion}
\noindent\paragraph{Comparison among algorithms} We compare the performance among PIE-rank, PIE-SS ( unweighted aggregation), CLeVer and Corona. First, we obtain the sensor selection results of them by using the training samples. Secondly, we calculate the distance matrix for each sensor crossing different samples (include both training and testing) by using DTW. Lastly, the final distance matrix $\textit{dist}_{unweighted}$ is calculated, and the classification performance is evaluated by NN-1 classifier. The accuracy results are reported in Fig.~\eqref{fig:fea_fs_mts}. The number of selected sensors is from 5 to 50 with step equals to 5. 

Overall, there is no single algorithm shows dominated higher accuracy. The performance of Corona and PIE-SS are quite close to each other and they show better performance than using all sensors (that means no sensor selection) generally. For CLeVer, its performance is not stable as it almost show worse classification accuracy than without doing any sensor selections in dataset ``DuckDuckGeese'' and ``EGG''. The PIE-rank algorithm show same unstable performance as PIE-SS. It shows lower performance in dataset ``DuckDuckGeese'' and ``EGG'', but has very good performance in other datasets. It even beats Corona and PIE-SS in dataset ``EGG'' and ``PEMS-SF'' at small number of selected sensors.

Meanwhile, we need to mention that ``CLever'' is an unsupervised sensor selection algorithm and the label information is not used for improving the classification performance. 

\noindent\paragraph{Compare the performance of without sensor selection} Another important evaluation is to measure the success of sensor selection. We draw a horizon line in each figure to show the performance of NN-1 classifier without any sensor selection. If one algorithm's curve is above the line, it means the sensor selection is success. Otherwise, it means fail. For example, in plot of ``DuckDuckGeese'', the PIE-rank and CLeVer algorithms are fail.

However, it worth to mention that PIE-rank show strong performance in dataset ``CMU-MOCAP-16'', ``KickvsPunch'', ``HumanGait'' and ``PEMS-SF''. Considering the advantage of its calculation time, it is quite an effective algorithm.

\noindent\paragraph{Comparison between weighted and unweighted aggregation} The result of PIE-SS by using different aggregation strategy is reported in Table~\eqref{tab:data_mts}. The weighted version show better classification performance than unweighted one except for dataset ``DuckDuckGeese''. The results match our expectation since we learn a linear combination of similarity graph to approximate the label matrix.

\section{Application to Heterogeneous Data}
\label{sec:heterogeneous}
In this section, we present an application of our PIE-rank algorithms on heterogeneous data. The motivation is to show the unique capability of our approach comparing to existing MTS related feature selection algorithms. Especially, (1) we avoid the feature extraction for time series, (2) we do not need the distance between two heterogeneous features.

We choose the public available data ``The PhysioNet Computing in Cardiology Challenge 2012''~\cite{silva2012predicting}~\cite{goldberger2000physiobank} for our experiment. The original goal of the challenge is to predict the mortality of ICU patients. Our goal is to rank the importance of features and compare the result with several published works~\cite{severeyn2012towards}~\cite{krajnak2012combining}~\cite{di2012robust}~\cite{mcmillan2012icu} in the Cardiology research domain. (We want to mention here that all the authors of this work have no domain knowledge about the Cardiology related research other than average common sense.) The introduced data is heterogeneous based on following observations:
\vspace{-10pt}
\begin{itemize}
    \setlength{\itemsep}{0pt}
    \setlength{\parskip}{0pt}
    \item Different type of feature values: real values, category values and time series, 
    \item The time series have different length not only among different sensor, also across different instances.
    \item The time series of different sensors have different sampling rate.
\end{itemize}

\begin{table}[h]
    \centering
    \begin{tabular}{|l|c|c|}%
    \toprule\
        Name & Set-A & Set-B \\ \hline
        \#Samples &  2099 & 2064 \\
        \#Label of ``1'' & 259 & 282 \\
        \#Real value features & 3 & 3 \\
        \#Category features & 2 & 2 \\
        \#MTS & 37 & 37 \\ \bottomrule
    \end{tabular}
    \caption{Summary of cleaned datasets of Physionet Challenge 2012. Label ``1'' means not survival.}
    \label{tab:summary_physionet_2012}
\end{table}
Except the heterogeneous characteristic, the data also include a lot of missing values. To prepossess the data, we follow part of the data cleaning procedures as introduced in work~\cite{silva2012predicting}~\cite{alistairewj}. To be specific, we only apply the steps in the author's Notebook before the section of feature extraction for time series. After prepossessing, we obtain two cleaned datasets ``Set-A'' and ``Set-B'' as in Table~\eqref{tab:summary_physionet_2012}.  When we build graph for each feature, we use Euclidean distance for feature: ``Age", ``Height" and ``Weight (static one)", and 1/0 distance for category feature: ``Gender" and ``ICUType". For time series features, we apply DTW distance. 

\begin{figure}[h]
    \centering
    \includegraphics[width=\columnwidth]{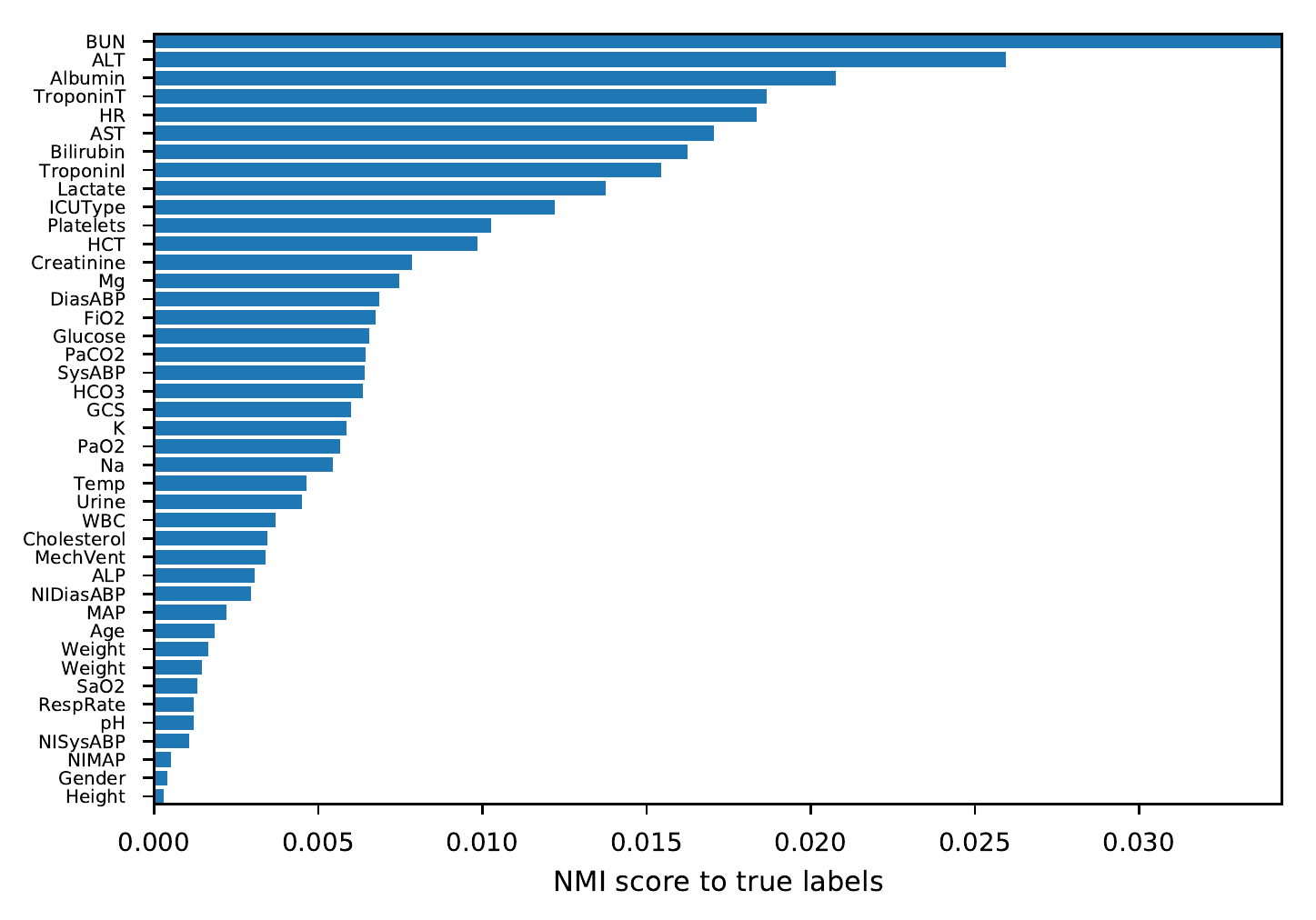}
    \caption{Ranking result of Physionet Challenge 2012 by PIE-rank.}
    \label{fig:physionet_rank_by_pie}
\end{figure}

We calculate the ranking scores of ``Set-A'' and ``Set-B'' by using ``PIE-rank'' and obtain two versions of ranking results. To obtain the final ranking scores, we use the average of them and report the result in Figure~\eqref{fig:physionet_rank_by_pie}.

\noindent\paragraph{Comparison to domain knowledge as in~\cite{krajnak2012combining}} In the work~\cite{krajnak2012combining}, the author provides 19 selected features by the clinicians (as shown in Table 2 of the original paper). Among, them 15 unique raw features are used. The other four repeat features with different extraction methods. We check the rank of these 15 raw features in our ranking result and see how many of them are ranked high. 
\begin{table}[h!]
    \centering
    \begin{tabular}{|l|c|}%
    \toprule
        Feature & Rank\\ \midrule
        Age &  33\\
        Bilirubin & 7\\
        BUN & 1\\ 
        Creatinine & 13\\
        Glasgow Coma Score (GCS) & 21\\
        HCO3 & 20 \\
        Heart Rate (HR) & 5\\
        PaO2 & 23\\
        pH & 38\\
        Platelets & 11\\
        Potassium (K) & 22\\
        Systolic ABP (SysABP) & 19\\
        Temp & 25\\
        Urine & 26\\
        White Blood Cell count (WBC) & 27\\ \bottomrule
    \end{tabular}
    \caption{Ranking result of 19 selected features by the clinicians in work~\cite{krajnak2012combining}.}
    \label{tab:rank_of_domain_knowledge}
\end{table}
From the result of Table~\eqref{tab:rank_of_domain_knowledge}, we found that 8 out of 15 raw features are ranked in the top half of original 42 raw features by our PIE-rank algorithm. 

\noindent\paragraph{Comparison to simple correspondence analysis as in~\cite{severeyn2012towards}}
The work ~\cite{severeyn2012towards} uses the simple correspondence analysis technique to analyze the importance of features that related to the survive or not. In the conclusion part (Section 4 of the original paper), the author list eight important features. The ranking result of them by PIE-rank are shown in table~\eqref{tab:ranking_of_sca_selected_feature}
\begin{table}[h!]
    \centering
    \begin{tabular}{|l|c|}
    \toprule
        Feature & Rank  \\ \hline
        Creatinine & 13\\
        Urine & 26\\
        BUN & 1\\
        Bilirubin & 7\\
        GCS & 21 \\
        MechVent & 29 \\  
        SOFA score & NA\\
        SAPS score & NA\\
        \bottomrule
    \end{tabular}
    \caption{Ranking result of eight selected features as in~\cite{severeyn2012towards}. The SOFA and SAPS scores are not included in our preprocessed data.}
    \label{tab:ranking_of_sca_selected_feature}
\end{table}
As we can see, 4 out of 6 selected features are ranked in the top half by our PIE-rank algorithm.

\noindent\paragraph{Comparison to feature selection result I as in~\cite{di2012robust}} In work~\cite{di2012robust}, the authors used forward sequential selection algorithm and cross-validation (CV) to select 32 most important features reported the occurrences of them in the repeated evaluations by CV. 
\begin{table}[h!]
    \centering
    \begin{tabular}{|l|l|}
    \toprule
        No & Features (rank by PIE-rank)  \\%
        \toprule
        \multirow{2}{*}{10} & Age(33), GCS(21), Temp(25),\\
                            & BUN(1), Glucose(17)\\ \hline%
        \multirow{2}{*}{9}  & Weight(34,35), Sodium(``Na"")(24),  \\
                            & WBC(27), Bilirubin(7), Cholesterol(28)\\ \hline%
        \multirow{2}{*}{8}  & Height(42), ICU Type(10),\\
                            & TroponinI(8), TroponinT(4)\\ \hline%
        \multirow{2}{*}{7}  & PaCO2(18), RespRate(37), HR(5), \\
                            & HCT(12), Albumin(3), ALP(30)\\ \hline%
        \multirow{2}{*}{6}  & SaO2(36), NISysABP(39), Mg(14), \\
                            & Platelets(11), Lactate(9)\\ \hline%
        5  & PaO2(23)\\ \hline%
        4  & SysABP(19)\\ \hline%
        \multirow{2}{*}{3}  & pH(38), MAP(32), NIDiasABP(31),\\
                            & ALT(2), AST(6)\\%
        \bottomrule
    \end{tabular}
    \caption{Ranking result of 32 selected features as in~\cite{di2012robust}. The ``Occurrence'' (the ``No'' column) values are provided in Table 1 of the cited work.}
    \label{tab:ranking_of_occurrence}
\end{table}

\noindent\paragraph{Comparison to feature selection result II as in~\cite{mcmillan2012icu}} In work~\cite{mcmillan2012icu}, the authors present three list of ranked top-10 features by their proposed algorithms. We compare their ranking results with us as in Table~\eqref{}.
\begin{table}[h!]
    \centering
    \begin{tabular}{|l|c|c|c|}
    \toprule
        Rank & Result-I & Result-2 & Result-3\\ 
        \midrule
        1 & K(22) & K(22) & K(22)\\
        2 & Age(33) & Albumin(3) & Temp(25)\\
        3 & ALP(30) & Age(33) & HR(5)\\
        4 & Platelets(11) & Glucose(17) & ALP(30)\\
        5 & Temp(25) & Urine(26) & Albumin(3) \\
        6 & Urine(26) & ALP(30) & PaO2(23)\\
        7 & Glucose(17) & GCS(21) & Age(33)\\
        8 & BUN(1) & pH(38) & Temp(25)\\
        9 & pH(38) & Albumin(3) & RespRate(37)\\
        10 & ALT(2) & Urine(26) & ALP(30)\\
        \bottomrule
    \end{tabular}
    \caption{Ranking results of selected features by~\cite{di2012robust}. The rank results are provided in Table 1 of the cited work. The value in the parenthesis are the ranking result of our PIE-rank algorithm.}
    \label{tab:ranking_of_icu}
\end{table}

\noindent\paragraph{Summary} From Table~\eqref{tab:rank_of_domain_knowledge} to \eqref{tab:ranking_of_icu}, we can see that different algorithms have different top-ranked features. Here we present a simple counting analysis of feature occurrences crossing four tables. We only list the features that appear at least three times out of four tables. The analysis results are shown in Table~\eqref{tab:ranking_summary}. 
\begin{table}[]
    \centering
    \begin{tabular}{|l|c|c|}
    \toprule
        No & Features & PIE-rank \\ \midrule
        4 & BUN & 1\\
        4 & GCS & 21\\
        3 & Age & 33\\
        3 & Urine & 26\\
        3 & HR & 5\\
        3 & PaO2 & 23\\
        3 & pH & 38\\
        3 & Platelets & 11\\
        3 & Temp & 25\\
        3 & Bilirubin & 7\\
        \bottomrule
    \end{tabular}
    \caption{Summary of ten features that are counted most among four tables. ``No'' means the occurrences.}
    \label{tab:ranking_summary}
\end{table}

There are ten features that appear at least three times among all four tables. Three of them are ranked in top-10 by our ``PIE-rank'' algorithm. The ``BUN'' and ``GCS'' features seems to be the most important features if we consider all results here (include our ranking result). The very contradictory features are ``Age'' and ``pH''that are considered important in Cardiology research field but show less impact by our algorithm regarding the survival or not for patients at ICU.
\section{Conclusion and Future Works}
\label{sec:conclusion}
In this work, we introduce two algorithms to rank and select subset of features (or sensors) for multivariate time series data. The unique part of our algorithms is that we avoid the feature extraction for each single time series which usually requires the domain knowledge regarding the input data. 

Our algorithms have substantial connection to multiple kernel learning. The adjacency matrix of our graph representation of each feature has the same form and meaning as a kernel matrix. It is natural to extend our work by connect it with kernel-based learning algorithms. Nevertheless, we also observe the computation issues when we build graph representation for each feature. In the future, we may try approximation algorithms such as Matrix sketching to reduce the computation time or apply deep learning algorithms to learn an embedding vector regarding the graph cluster structure directly.





\bibliography{tfs}
\end{document}